\documentclass{article}
\usepackage{times}
\usepackage{graphicx}
\usepackage{algorithm,algorithmic,a4wide,amssymb}
\begin{document}

\title{My First Deep Learning System of 1991 \\
 $+$ Deep Learning Timeline 1962-2013 }

\date{20 December 2013}
\author{J\"{u}rgen Schmidhuber \\
The Swiss AI Lab IDSIA, Galleria 2, 6928 Manno-Lugano \\
University of Lugano \& SUPSI, Switzerland}
\maketitle

\begin{abstract}
Deep Learning has attracted significant attention in recent years.
Here I present a brief overview of my first Deep Learner of 1991, and its historic context,
with a timeline of Deep Learning highlights.

{\em Note:}
As a machine learning researcher I am obsessed with proper credit assignment.
This draft is the result of an experiment in rapid massive open online peer review.
Since 20 September 2013, subsequent revisions published under {\em www.deeplearning.me} have
 absorbed many suggestions for improvements by experts.
The abbreviation ``TL" is used to refer to subsections of the timeline section.
\end{abstract}


\begin{figure}[htb]
\begin{center}
\includegraphics[width=\linewidth]{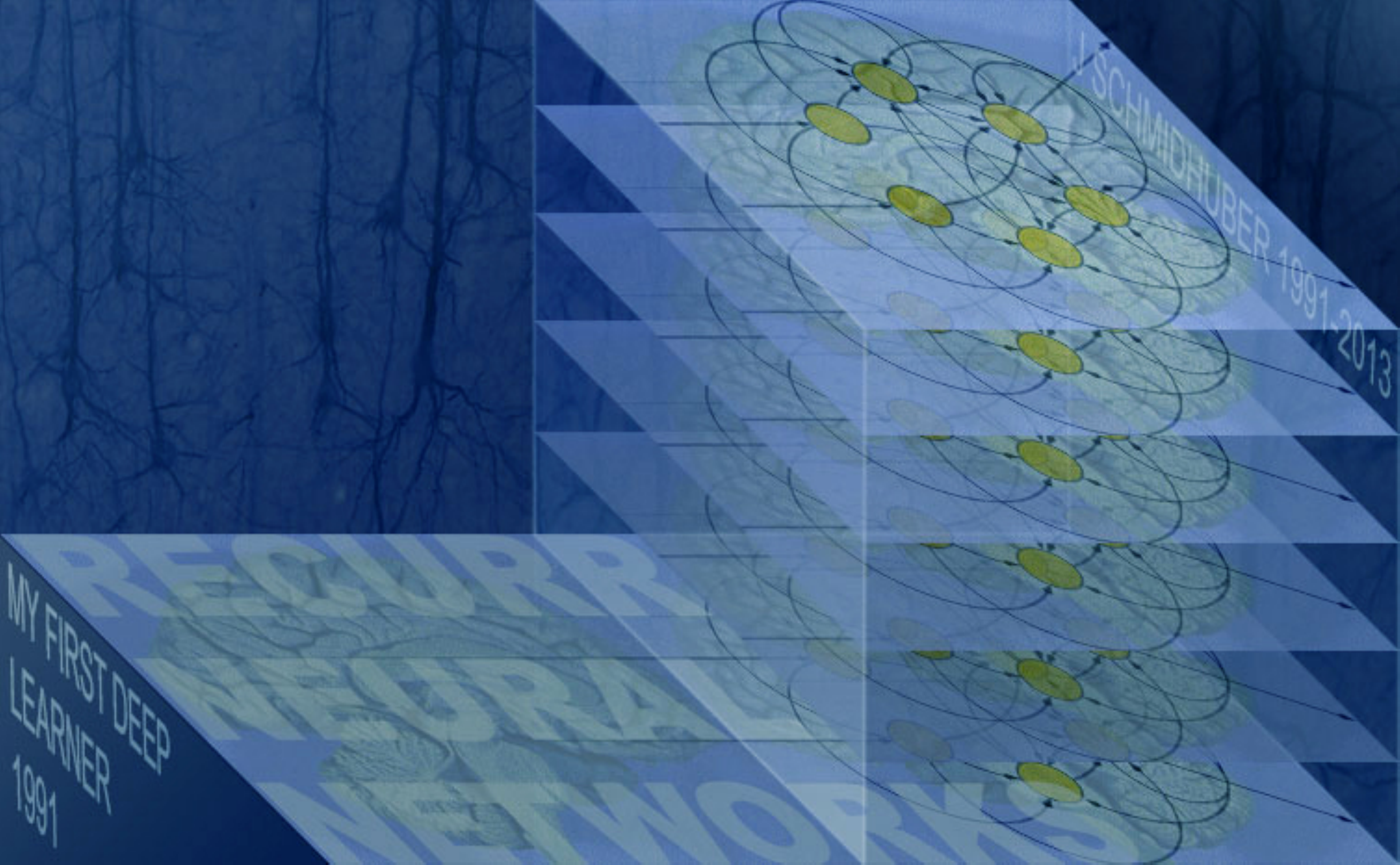}
\end{center}
\caption{My first Deep Learning system of 1991 used   
a deep stack of recurrent neural networks (a {\em Neural Hierarchical Temporal Memory}) pre-trained in unsupervised fashion 
to accelerate subsequent supervised learning 
\cite{Schmidhuber:91chunker,Schmidhuber:92ncchunker,Schmidhuber:93habil}.}
\label{fig:}
\end{figure}

\newpage

In 2009, our Deep Learning Artificial Neural Networks became the first Deep Learners to win official international pattern recognition competitions~\cite{graves:2009nips,schmidhuber2011agi} (with secret test sets known only to the organisers); by 2012 they had won eight of them (TL \ref{2012}), including the first contests on object 
detection in large images (ICPR 2012) \cite{icpr12,miccai2013}  and image segmentation (ISBI 2012) \cite{isbi12,ciresan2012nips}.
 In 2011, they achieved the world's first superhuman visual pattern recognition results~\cite{ciresan:2012NN,ciresan:2011ijcnn}. Others have implemented 
very similar techniques, e.g., \cite{Krizhevsky:2012}, and won additional contests or set
benchmark records since 2012, e.g., (TL \ref{2012}, TL \ref{2013}). The field of Deep Learning research is far older though---compare the timeline (TL) further down.

My first Deep Learner dates back to 1991~\cite{Schmidhuber:91chunker,Schmidhuber:92ncchunker,Schmidhuber:93habil} (TL \ref{1991b}). It can perform credit assignment across hundreds of nonlinear operators or neural layers, by using unsupervised pre-training for a stack of recurrent neural networks (RNN) (deep by nature) as in Figure 1. Such RNN are general computers more powerful than normal feedforward NN, and can encode entire sequences of input vectors.

The basic idea is still relevant today. Each RNN is trained for a while in unsupervised fashion to predict its next input. From then on, only unexpected inputs (errors) convey new information and get fed to the next higher RNN which thus ticks on a slower, self-organising time scale. It can easily be shown that no information gets lost. It just gets compressed (much of machine learning is essentially about compression). We get less and less redundant input sequence encodings in deeper and deeper levels of this {\em hierarchical temporal memory}, which compresses data in both space (like feedforward NN) and time. There also is a continuous variant~\cite{SchmidhuberMozerPrelinger:93}.

One ancient illustrative Deep Learning experiment of 1993~\cite{Schmidhuber:93habil} required credit assignment across 1200 time steps, or through 1200 subsequent nonlinear virtual layers. The top level code of the initially unsupervised RNN stack, however, got so compact that (previously infeasible) sequence classification through additional supervised learning became possible.

There is a way of compressing higher levels down into lower levels, thus partially collapsing the hierarchical temporal memory. The trick is to retrain lower-level RNN to continually imitate (predict) the hidden units of already trained, slower, higher-level RNN, through additional predictive output neurons~\cite{Schmidhuber:92ncchunker,Schmidhuber:91chunker,Schmidhuber:93habil}. This helps the lower RNN to develop appropriate, rarely changing memories that may bridge very long time lags.

The Deep Learner of 1991 was a first way of overcoming the {\em Fundamental Deep Learning Problem} identified and analysed in 1991 by my very first student (now professor) Sepp Hochreiter (TL \ref{1991a}): the problem of vanishing or exploding gradients~\cite{Hochreiter:91,Hochreiter:01book,Bengio:94}. The latter motivated all our subsequent Deep Learning research of the 1990s and 2000s.

Through supervised LSTM RNN (1997) (e.g., \cite{Hochreiter:97lstm,Gers:99c,graves05nn,Graves:06icml,Graves:09tpami,graves:2009nips,graves:2013icassp,schmidhuber2011agi}, TL \ref{1997}) we could eventually perform similar feats as with the 1991 system~\cite{Schmidhuber:92ncchunker,Schmidhuber:93habil}, overcoming the {\em Fundamental Deep Learning Problem} without any unsupervised pre-training. Moreover, LSTM could also learn tasks unlearnable by the partially unsupervised 1991 chunker~\cite{Schmidhuber:92ncchunker,Schmidhuber:93habil}.

Particularly successful are stacks of LSTM RNN~\cite{graves:2009nips} trained by {\em Connectionist Temporal Classification} (CTC)~\cite{Graves:06icml}. On faster computers of 2009, this became the first RNN system ever to win an official international pattern recognition competition~\cite{graves:2009nips,schmidhuber2011agi}, through the work of my PhD student and postdoc Alex Graves, e.g., \cite{graves:2009nips}. To my knowledge, this also was the first Deep Learning system ever (recurrent or not) to win such a contest (TL \ref{2009}). (In fact, it won three different ICDAR 2009 contests on connected handwriting in three different languages, e.g., \cite{schmidhuber2011agi,graves:2009nips}, TL \ref{2009}.) A while ago, Alex moved on to Geoffrey Hinton's lab (Univ. Toronto), where a stack~\cite{graves:2009nips} of our bidirectional LSTM RNN~\cite{graves05nn} also broke a famous TIMIT speech recognition record~\cite{graves:2013icassp} (TL \ref{2013}), despite thousands of man years previously spent on HMM-based speech recognition research. CTC-LSTM also 
helped to score first at NIST's OpenHaRT2013 evaluation ~\cite{bluche13}.

Recently, well-known entrepreneurs also got interested \cite{hawkins2006,kurzweil2012} in 
  such hierarchical temporal memories ~\cite{Schmidhuber:92ncchunker,Schmidhuber:93habil} (TL \ref{1991b}).

The expression {\em Deep Learning} actually got coined relatively late, around 2006, in the context of unsupervised pre-training for less general feedforward networks~\cite{HinSal06} (TL \ref{2006}). Such a system reached 1.2\% error rate~\cite{HinSal06} on the MNIST handwritten digits~\cite{LeCun:89,LeCun:90}, perhaps the most famous benchmark of Machine Learning. Our team first showed that good old backpropagation (TL \ref{1970}) on GPUs (with training pattern distortions~\cite{Baird90,simard:2003} but without any unsupervised pre-training) can actually achieve a three times better result of 0.35\%~\cite{ciresan:2010} - back then, a world record (a previous standard net achieved 0.7\%~\cite{simard:2003}; a backprop-trained~\cite{LeCun:89,LeCun:90} {\em Convolutional NN (CNN or convnet)}~\cite{Fukushima:1979neocognitron,fukushima:1980,LeCun:89,LeCun:90} got 0.39\%~\cite{ranzato-06}(TL \ref{2006}); plain backprop without distortions except for small saccadic eye movement-like translations already got 0.95\%). Then we replaced our standard net by 
a biologically rather plausible architecture inspired by early neuroscience-related work~\cite{Fukushima:1979neocognitron,Hubel:62,fukushima:1980,LeCun:89}: {\em Deep and Wide GPU-based Multi-Column Max-Pooling CNN (MCMPCNN)}~\cite{ciresan:2011ijcai,ciresan:2012,ciresan2012cvpr} with alternating backprop-based~\cite{LeCun:89,LeCun:90,ranzato-cvpr-07} weight-sharing convolutional layers~\cite{fukushima:1980,LeCun:89,lecun:1998,Behnke:LNCS} and winner-take-all~\cite{Fukushima:1979neocognitron,fukushima:1980} max-pooling~\cite{riesenhuber:1999,ranzato-cvpr-07,scherer:2010} layers (see~\cite{chellapilla:2006b} for early GPU-based CNN). MCMPCNN are committees of MPCNN~\cite{ciresan:2011ijcnn} with simple democratic output averaging (compare earlier more sophisticated ensemble methods~\cite{Schapire:90}). Object detection~\cite{miccai2013}  and image segmentation~\cite{ciresan2012nips}  
(TL \ref{2012})
profit from fast MPCNN-based image scans~\cite{masci:2013icip,giusti:2013icip}. Our supervised GPU-MCMPCNN was the 
first system to achieve superhuman performance in an official international competition (with deadline and secret test set known only to the organisers)~\cite{ciresan:2012NN,ciresan:2011ijcnn} (TL \ref{2011}) (compare~\cite{sermanet-ijcnn-11}), and the first with human-competitive performance (around 0.2\%) on MNIST~\cite{ciresan2012cvpr}. Since 2011, it has won numerous additional competitions on a routine basis, e.g.,  (TL \ref{2012}, TL \ref{2013}).

Our (multi-column \cite{ciresan:2011ijcnn})  GPU-MPCNN \cite{ciresan:2011ijcai} (TL \ref{2011})
 were adopted by the groups of {\em Univ. Toronto / Stanford / Google,} e.g., \cite{Krizhevsky:2012,coates:2013icml} (TL \ref{2012}, TL \ref{2013}). {\em Apple Inc.,} the most profitable smartphone maker, hired Ueli Meier, member of our Deep Learning team that won the ICDAR 2011 Chinese handwriting contest~\cite{ciresan2012cvpr}. {\em ArcelorMittal,} the world's top steel producer, is using our methods for material defect detection and classification, e.g., \cite{masci:2013icip}. Other users include a leading automotive supplier, recent start-ups such as {\em deepmind} (which hired four of my former PhD students/postdocs), and many other companies and leading research labs. One of the most important applications of our techniques is biomedical imaging ~\cite{miccai2013} (TL \ref{2012}, TL \ref{2013}), e.g., for cancer prognosis or plaque detection in CT heart scans.

Remarkably, the most successful Deep Learning algorithms in most international contests since 2009 (TL \ref{2009}-\ref{2013}) are adaptations and extensions of an over 40-year-old algorithm, namely, supervised efficient backprop~\cite{Linnainmaa:1970,Werbos:81sensitivity} (TL \ref{1970}) (compare~\cite{Linnainmaa:1976,LeCun:85,Rumelhart:86,bryson1969applied,Parker:85}) or BPTT/RTRL for RNN, e.g., \cite{Williams:89,RobinsonFallside:87tr,Werbos:88gasmarket,pear89,Schmidhuber:92ncn3,Martens:2011hessfree}  
(exceptions include two 2011 contests specialised on transfer learning~\cite{goodfellow:2012icml}---but compare~\cite{Ciresan:2012a}). 
In particular, as of 2013, state-of-the-art {\em feedforward} nets (TL \ref{2011}-\ref{2012}) are GPU-based~\cite{ciresan:2011ijcai} multi-column~\cite{ciresan2012cvpr} combinations of two ancient concepts: Backpropagation (TL \ref{1970}) applied~\cite{LeCun:90} to Neocognitron-like convolutional 
architectures (TL \ref{1979}) (with max-pooling layers~\cite{riesenhuber:1999,ranzato-cvpr-07,scherer:2010} instead of alternative~\cite{Fukushima:1979neocognitron,fukushima:1980,Fukushima:2013,Schmidhuber:89cs} local winner-take-all methods). (Plus additional tricks from the 1990s and 2000s, e.g., \cite{orr1998neural,tricksofthetrade:2012}.) 
In the quite different deep {\em recurrent} case, supervised systems also dominate, e.g.,  \cite{Hochreiter:97lstm,Graves:06icml,graves:2009nips,Graves:09tpami,Martens:2011hessfree,graves:2013icassp} (TL \ref{2009}, TL \ref{2013}).

In particular, most competition-winning or benchmark record-setting Deep Learners (TL \ref{2009} - TL \ref{2013}) use one of two {\em supervised} techniques developed in my lab: (1) recurrent LSTM (1997) (TL \ref{1997}) trained by CTC (2006) \cite{Graves:06icml}, or (2) feedforward GPU-MPCNN (2011) \cite{ciresan:2011ijcai} (TL \ref{2011}) (building on earlier work since the 1960s mentioned in the text above). 
Nevertheless, in many applications it can still be advantageous to combine the best of both worlds - {\em supervised} learning and {\em unsupervised} pre-training, like in my 1991 system described above~\cite{Schmidhuber:91chunker,Schmidhuber:92ncchunker,Schmidhuber:93habil}.


\newpage

\section{Timeline of Deep Learning Highlights}
\label{timeline}

\subsection{1962: Neurobiological Inspiration Through Simple Cells and Complex Cells}
\label{1962}
Hubel and Wiesel described simple cells and complex cells in the visual cortex~\cite{Hubel:62}.
This  inspired later deep artificial neural network (NN) architectures 
(TL \ref{1979}) used in certain modern award-winning Deep Learners (TL \ref{2011}-\ref{2013}).
(The author of the present paper was conceived in 1962.)

\subsection{1970 $\pm$ a Decade or so: Backpropagation}
\label{1970}
Error functions and their gradients for complex, 
nonlinear, multi-stage, differentiable, NN-related systems have been discussed 
at least since the early 1960s, e.g., \cite{Griewank:2012,Kelley:1960,bryson:1961,dreyfus:1962,bryson1969applied,Wilkinson:1988,Amari:1967:TAP,director:1969}. 
Gradient descent~\cite{hadamard1908memoire} in such systems can be performed \cite{bryson:1961,Kelley:1960,bryson1969applied} 
by iterating the ancient chain rule~\cite{leibniz:1676,de1716analyse} in dynamic programming style~\cite{Bellman:1957} 
(compare simplified derivation using chain rule only~\cite{dreyfus:1962}). 
However, efficient error backpropagation (BP) in arbitrary, possibly sparse, 
NN-like networks apparently was first described by 
Linnainmaa in 1970~\cite{Linnainmaa:1970,Linnainmaa:1976} (he did not refer to NN though).
BP is also known as the reverse mode of automatic differentiation~\cite{Griewank:2012}, 
where the costs of forward activation spreading essentially equal the costs of backward 
derivative calculation. See early FORTRAN code~\cite{Linnainmaa:1970}, and compare~\cite{ostrovskii:1971}.
Compare the concept of ordered derivative~\cite{Werbos:74} and related work \cite{dreyfus:1973}, with NN-specific discussion~\cite{Werbos:74} 
(section 5.5.1), and the first NN-specific efficient BP of 1981 by Werbos~\cite{Werbos:81sensitivity,werbos2006backwards}.
Compare \cite{LeCun:85,Rumelhart:86,Parker:85} and generalisations for sequence-processing 
recurrent NN, e.g., \cite{Williams:89,RobinsonFallside:87tr,Werbos:88gasmarket,pear89,Schmidhuber:92ncn3,Martens:2011hessfree}. 
See also natural gradients~\cite{amari1998natural}. 
As of 2013, BP is still the central Deep Learning algorithm.

\subsection{1979: Deep Neocognitron, Weight Sharing, Convolution}
\label{1979}
Fukushima's deep Neocognitron architecture \cite{Fukushima:1979neocognitron,fukushima:1980,Fukushima:2013}
incorporated neurophysiological insights (TL \ref{1962}) ~\cite{Hubel:62}. It introduced weight-sharing 
{\em Convolutional Neural Networks} (CNN) as well as winner-take-all layers. It is
very similar to the architecture of modern, competition-winning, purely {\em  supervised}, 
feedforward, gradient-based Deep Learners (TL \ref{2011}-\ref{2013}).
Fukushima, however, used local {\em un}supervised learning rules instead.

\subsection{1987: Autoencoder Hierarchies}
\label{1987}
In 1987, Ballard published ideas on unsupervised autoencoder hierarchies~\cite{ballard1987modular}, 
related to post-2000 feedforward Deep Learners (TL \ref{2006}) based on unsupervised pre-training, 
e.g., \cite{HinSal06}; compare survey~\cite{hinton1989connectionist} and somewhat 
related RAAMs~\cite{pollack1988implications}.

\subsection{1989: Backpropagation for CNN}
\label{1989}
LeCun {\em et al.} \cite{LeCun:89,LeCun:90} applied
backpropagation (TL \ref{1970}) to Fukushima's weight-sharing convolutional 
neural layers (TL \ref{1979})~\cite{Fukushima:1979neocognitron,fukushima:1980,LeCun:89}.
This combination has become an 
essential ingredient of many modern, competition-winning, feedforward, visual Deep Learners (TL \ref{2011}-\ref{2012}).

\subsection{1991: Fundamental Deep Learning Problem}
\label{1991a}
By the early 1990s, experiments had shown that deep feedforward or recurrent networks are hard to
train by backpropagation (TL \ref{1970}). My student
Hochreiter~\cite{Hochreiter:91} discovered and analyzed the reason, namely, the
{\em Fundamental Deep Learning Problem} due to vanishing or exploding gradients. 
Compare~\cite{Hochreiter:01book}.

\subsection{1991: Deep Hierarchy of Recurrent NN}
\label{1991b}
My first recurrent Deep Learning system (present paper)  
partially overcame the fundamental problem (TL \ref{1991a})
through a deep RNN stack pre-trained in unsupervised fashion 
\cite{Schmidhuber:91chunker,Schmidhuber:92ncchunker,Schmidhuber:93habil}
to accelerate subsequent supervised learning.
This was a working Deep Learner in the 
modern post-2000 sense, and also the first {\em Neural Hierarchical Temporal Memory}.

\subsection{1997: Supervised Deep Learner (LSTM)}
\label{1997}
Long Short-Term Memory (LSTM) recurrent neural networks (RNN)
became the
first purely supervised Deep Learners, e.g., 
\cite{Hochreiter:97lstm,Gers:2000nc,graves05nn,Graves:06icml,Graves:09tpami,graves:2009nips,graves:2013icassp}. 
LSTM RNN were able to learn solutions to many previously unlearnable problems (see also TL \ref{2009}, TL \ref{2013}).

\subsection{2006: Deep Belief Networks / CNN Results}
\label{2006}
A paper by Hinton and Salakhutdinov~\cite{HinSal06} focused on 
unsupervised pre-training of feedforward NN 
to accelerate subsequent supervised learning (compare TL \ref{1991b}).
This helped to arouse interest in deep NN (keywords: restricted Boltzmann machines;
Deep Belief Networks).
 In the same year, a BP-trained CNN (TL \ref{1979}, TL \ref{1989}) 
by Ranzato {\em et al.} \cite{ranzato-06}
set a new record   on
the famous MNIST handwritten digit recognition benchmark~\cite{LeCun:89},
using training pattern
deformations~\cite{Baird90,simard:2003}.

\subsection{2009: First Competitions Won by Deep Learning}
\label{2009}
2009 saw the first Deep Learning systems to win
official international pattern recognition contests (with secret test sets known
only to the organisers):
three connected handwriting competitions at ICDAR 2009 were won by deep
LSTM RNN~\cite{graves:2009nips,schmidhuber2011agi}
performing simultaneous segmentation and recognition.

\subsection{2010: Plain Backpropagation on GPUs Yields Excellent Results}
\label{2010}
 In 2010, a new MNIST record was set 
by good old backpropagation (TL \ref{1970}) in deep but otherwise
standard NN, without unsupervised pre-training, and without convolution (but with training pattern
deformations). 
This was made possible mainly by boosting computing power through a fast GPU implementation
\cite{ciresan:2010}. (A year later, first human-competitive performance on MNIST
was achieved by a deep MCMPCNN (TL \ref{2011}) \cite{ciresan2012cvpr}.)

\subsection{2011: MPCNN on GPU / First Superhuman Visual Pattern Recognition}
\label{2011}
In 2011, Ciresan {\em et al.} introduced supervised {\em GPU-based Max-Pooling CNN or Convnets (MPCNN)} \cite{ciresan:2011ijcai},
today used by most if not all feedforward competition-winning deep NN (TL \ref{2012}, TL \ref{2013}).
The first superhuman visual pattern recognition in a controlled competition
 (traffic signs ~\cite{traffic11}) was achieved ~\cite{ciresan:2012NN,ciresan:2011ijcnn}  (twice better than humans, three times better than the closest artificial NN competitor, 
six times better than the best non-neural method),
through {\em deep and wide Multi-Column (MC) GPU-MPCNN} \cite{ciresan:2011ijcai,ciresan:2011ijcnn}, the current gold standard for deep feedforward NN.

\subsection{2012: First Contests Won on Object Detection and Image Segmentation}
\label{2012}
2012 saw the first Deep learning system (a GPU-MCMPCNN \cite{ciresan:2011ijcai,ciresan:2011ijcnn}, TL \ref{2011}) to win a 
contest on {\em visual object  detection} in large images 
(as opposed to mere {\em recognition/classification}):
the {\em ICPR 2012 Contest on Mitosis Detection in Breast Cancer Histological Images}
~\cite{icpr12,icpr12report,miccai2013}. 
An MC (TL \ref{2011}) variant of a GPU-MPCNN also achieved
best results on the {\em ImageNet} classification benchmark  \cite{Krizhevsky:2012}.
2012 also saw the 
first pure image {\em segmentation} contest won by Deep Learning (again through a GPU-MCMPCNN),
namely, the
{\em ISBI 2012 Challenge on segmenting neuronal structures}
 ~\cite{isbi12,ciresan2012nips}. This was the
8th international pattern recognition contest won
by my team since 2009~\cite{interview2012}.

\subsection{2013: More Contests and Benchmark Records}
\label{2013}
In 2013, a new {\em TIMIT phoneme recognition} record 
was set by deep LSTM RNN ~\cite{graves:2013icassp} (TL \ref{1997}, TL \ref{2009}).
A new record \cite{chinese2013} on the {\em ICDAR Chinese handwriting recognition 
benchmark} (over 3700 classes) was set on a desktop machine
by a GPU-MCMPCNN with almost human performance. 
The {\em MICCAI 2013 Grand Challenge on Mitosis Detection}
was won by a GPU-MCMPCNN ~\cite{miccai13,miccai2013}.
Deep GPU-MPCNN \cite{ciresan:2011ijcai} also helped to achieve new best results
on ImageNet classification \cite{zeiler2013} and PASCAL object detection \cite{malik2013}.
Additional contests are mentioned in the web pages of
the Swiss AI Lab IDSIA,
the University of Toronto,
NY University,
and the University of Montreal.

\section{Acknowledgments}
\label{ack}  
Drafts/revisions of this paper have been published since 20 Sept 2013 in my massive open peer review web site
{\em  www.idsia.ch/\~{}juergen/firstdeeplearner.html} (also under {\em www.deeplearning.me}).
Thanks for valuable comments 
 to Geoffrey Hinton, Kunihiko Fukushima, Yoshua Bengio, Sven Behnke, Yann LeCun, Sepp Hochreiter, Mike Mozer, Marc'Aurelio Ranzato, Andreas Griewank, Paul Werbos, Shun-ichi Amari, Seppo Linnainmaa, Peter Norvig, Yu-Chi Ho, Alex Graves, Dan Ciresan, Jonathan Masci, Stuart Dreyfus, and others. 

\bibliography{bib}
\bibliographystyle{abbrv}
\end{document}